\documentclass[11pt,a4paper]{article}
\usepackage[hyperref]{eacl2021}
\usepackage{times}
\usepackage{latexsym}

\usepackage{microtype}

\usepackage{tabularx}
\usepackage{booktabs}
\usepackage{makecell}
\usepackage{multirow}
\usepackage{amssymb}
\usepackage{amsmath}
\usepackage{subcaption}
\usepackage{xspace}
\usepackage{bm}
\usepackage{appendix}

\usepackage{paralist}

\usepackage{algorithm}
\usepackage[noend]{algpseudocode}
\usepackage{graphicx}

\newcolumntype{Y}{>{\centering\arraybackslash}X}

\aclfinalcopy 
\def\aclpaperid{495} 

\newcommand{\pn}{\hphantom{-}}

\newcommand{\crafty}{Crafty\xspace}

\title{On the Evaluation of Vision-and-Language Navigation Instructions}

\author{Ming Zhao \quad Peter Anderson \quad Vihan Jain \quad Su Wang\\ \quad \textbf{Alexander Ku}  \quad \textbf{Jason Baldridge} \quad \textbf{Eugene Ie}\\
Google Research\\
\texttt{\{astroming, pjand, vihan, wangsu, alexku, jridge, eugeneie\}}\\ 
\texttt{@google.com}}

\date{}

\begin{document}
\maketitle

\begin{abstract}
Vision-and-Language Navigation wayfinding agents can be enhanced by exploiting automatically generated navigation instructions. However, existing instruction generators have not been comprehensively evaluated, and the automatic evaluation metrics used to develop them have not been validated. Using human wayfinders, we show that these generators perform on par with or only slightly better than a template-based generator and far worse than human instructors. Furthermore,  we discover that BLEU, ROUGE, METEOR and CIDEr are ineffective for evaluating grounded navigation instructions.
To improve instruction evaluation, we propose an instruction-trajectory compatibility model that operates without reference instructions. Our model shows the highest correlation with human wayfinding outcomes when scoring individual instructions. For ranking instruction generation systems, if reference instructions are available we recommend using SPICE.
\end{abstract}

\section{Introduction}
\label{sec:intro}

Generating route instructions is a long studied problem with clear practical applications \cite{richter:klippel:2005}. Whereas earlier work sought to create instructions for human wayfinders, recent work has focused on using instruction-generation models to improve the performance of agents that follow instructions given by people. In the context of Vision-and-Language Navigation (VLN) datasets such as Room-to-Room (R2R) \cite{mattersim}, models for generating navigation instructions have improved agents' wayfinding performance in at least two ways: (1) by synthesizing new instructions for data augmentation \cite{fried2018speaker, backtranslate2019}, and (2) by fulfilling the role of a probabilistic speaker in a pragmatic reasoning setting \cite{fried2018speaker}. Such data augmentation is so effective that it is nearly ubiquitous in the best performing agents \cite{wang2018reinforced,Huang2019Transferable,li2019robust}. 

To make further advances in the generation of visually-grounded navigation instructions, accurate evaluation of the generated text is essential. However, the performance of existing instruction generators has not yet been evaluated using human wayfinders, and the efficacy of the automated evaluation metrics used to develop them has not been established. This paper addresses both gaps.

\begin{figure}
\centering
\scriptsize
\includegraphics[width=0.49\textwidth]{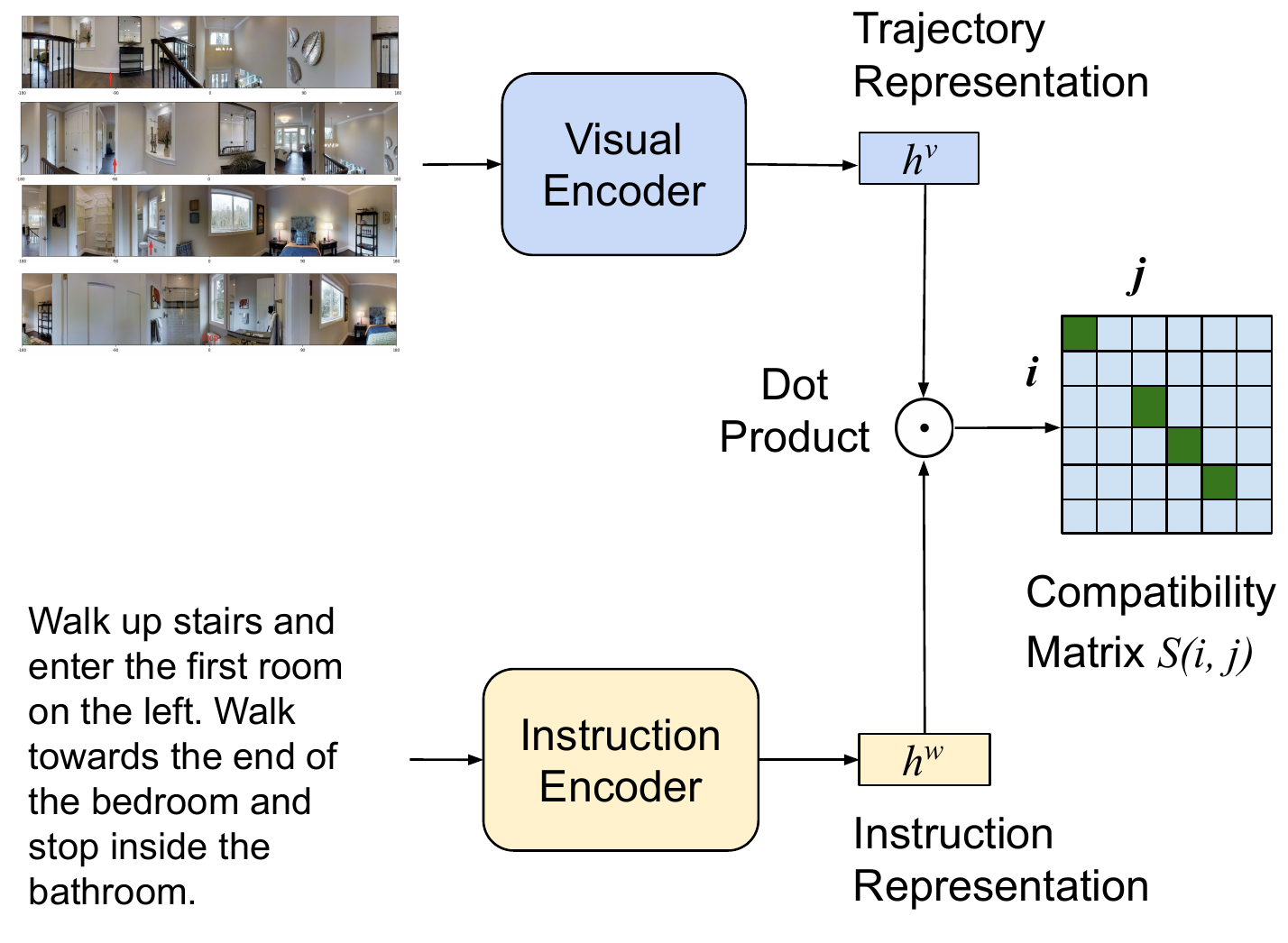}
\caption{Proposed dual encoder instruction-trajectory compatibility model. Navigation instructions and trajectories (sequences of panoramic images and view angles) are projected into a shared latent space. The independence between the encoders facilitates learning using both contrastive and classification losses.}
\label{fig:architecture}
\end{figure}

To establish benchmarks for navigation instruction generation, we evaluate existing English models \cite{fried2018speaker, backtranslate2019} using human wayfinders. These models are effective for data augmentation, but in human trials they perform on par with or only slightly better than a template-based system, and they are far worse than human instructors. This leaves much headroom for better instruction generation, which may in turn improve agents' wayfinding abilities.

Next, we consider the evaluation of navigation instructions without human wayfinders, a necessary step for future improvements in both grounded instruction generation (itself a challenging and important language generation problem) and agent wayfinding. We propose a model-based approach (Fig. \ref{fig:architecture}) to measure the compatibility of an instruction-trajectory pair without needing reference instructions for evaluation. In training this model, we find that adding contrastive losses in addition to pairwise classification losses improves AUC by 9--10\%, round-trip back-translation improves performance when used to paraphrase positive examples, and that both trajectory and instruction perturbations are useful as hard negatives.

Finally, we compare our compatibility model to common textual evaluation metrics to assess which metric best correlates with the outcomes of human wayfinding attempts. We discover that BLEU \cite{Papineni2002}, ROUGE \cite{Lin2004}, METEOR \cite{denkowski2014} and CIDEr \cite{Vedantam2015} are ineffective for evaluating grounded navigation instructions. For system-level evaluations with reference instructions, we recommend SPICE \cite{spice2016}.  When averaged over many instructions, SPICE correlates with both human wayfinding performance and subjective human judgments of instruction quality. When scoring individual instructions, our compatibility model most closely reflects human wayfinding performance, outperforming BERTScore \citep{Zhang2019bertscore} and VLN agent-based scores. Our results are a timely reminder that textual evaluation metrics should always be validated against human judgments when applied to new domains. We plan to release our trained compatibility model and the instructions and human evaluation data we collected.

\section{Related Work}
\label{sec:related_work}

\paragraph{Navigation Instruction Generation}
Until recently, most methods for generating navigation instructions were focused on settings in which a system has access to a map representation of the environment, including the locations of objects and named items (e.g. of streets and buildings) \cite{richter:klippel:2005}. Some generate route instructions \textit{interactively} given the current position and goal location \cite{drager-koller-2012-generation}, while others provide \textit{in-advance} instructions that must be more robust to possible misinterpretation \cite{roth-frank-2010-computing,mast:wolter:2013}.

Recent work has focused on instruction generation to improve the performance of wayfinding agents. Two instruction generators, \textit{Speaker-Follower} \cite{fried2018speaker} and \textit{EnvDrop} \cite{backtranslate2019}, have been widely used for R2R data augmentation. They provide $\sim$170k new instruction-trajectory pairs sampled from training environments. Both are seq-to-seq models with attention. They take as input a sequence of panoramas grounded in a 3D trajectory, and output a textual instruction intended to describe it.

\paragraph{Vision-and-Language Navigation}
For VLN, embodied agents in 3D environments must follow natural language instructions to reach prescribed goals. Most recent efforts \citep[e.g.,][etc.]{Fu2019CounterfactualVLN, Huang2019Transferable, jain2019stayonpath,  wang2018reinforced} have used the Room-to-Room (R2R) dataset \citep{mattersim}, which contains 4675 unique paths in the \textit{train} split, 340 in the \textit{val-seen} split (same environments, new paths), and an additional 783 paths in the \textit{val-unseen} split (new environments, new paths). However, our findings are also relevant for similar datasets such as Touchdown \cite{chen2019touchdown,mehta2020retouchdown}, CVDN \cite{thomason:corl19}, REVERIE \cite{qi2019reverie}, and the multilingual Room-across-Room (RxR) dataset \cite{ku-etal-2020-room}.

\paragraph{Text Generation Metrics}
There are many automated metrics that assess textual similarity; we focus on five that are extensively used in the context of image captioning: BLEU \citep{Papineni2002}, METEOR \cite{denkowski2014}, ROUGE \cite{Lin2004}, CIDEr \cite{Vedantam2015} and SPICE \cite{spice2016}. 
More recently, model- and semi-model-based metrics have been proposed. BERTScore \citep{Zhang2019bertscore} takes a semi-model-based approach to compute token-wise similarity using contextual embeddings learned with BERT \cite{devlin2018bert}. 
BLEURT \citep{Sellam2020bleurt} is a fully model-based approach  combining large-scale synthetic pretraining and domain specific finetuning. 
However, all of the aforementioned metrics are reference-based, and none is specifically designed for assessing navigation instructions associated with 3D trajectories for an embodied agent, which requires not only language-to-vision grounding but also correct sequencing.

\paragraph{Instruction-Trajectory Compatibility Models} 
Our model builds on that of \citet{Huang2019Transferable}, but differs in loss (using focal and contrastive losses), input features (adding action and geometry representation), and negative mining strategies (adding instruction perturbations in addition to trajectory perturbations). Compared to the trajectory re-ranking compatibility model proposed by \citet{Majumdar2020improving}, we use a dual encoder architecture rather than dense cross-attention. This facilitates the efficient computation of contrastive losses, which are calculated over all pairs in a minibatch, and improve AUC by 10\% in our model. We also avoid training on the outputs of the instruction generators (to prevent overfitting to the models we evaluate). We are yet to explore transfer learning (which is the focus of \citet{Majumdar2020improving}).  

\section{Human Wayfinding Evaluations}
\label{sec:human_eval}

To benchmark the current state-of-the-art for navigation instruction generation, we evaluate the outputs of the \textit{Speaker-Follower} and \textit{EnvDrop} models by asking people to follow them. We use instructions for the 340 and 783 trajectories in the R2R val-seen and val-unseen splits, respectively. Both models are trained on the R2R train split and the generated instructions were provided by the respective authors. To contextualize the results, we additionally evaluate instructions from a template-based generator (using ground-truth object annotations), a new set of instructions written by human annotators, and three adversarial perturbations of these human instructions. 
New navigation instructions and wayfinding evaluations are collected using a lightly modified version of PanGEA\footnote{\href{https://github.com/google-research/pangea}{https://github.com/google-research/pangea}}, an open-source annotation toolkit for panoramic graph environments.

\paragraph{\crafty} \crafty is a template-based navigation instruction generator. It observes the trajectory's geometry and nearby ground-truth object annotations, identifies salient objects, and creates English instructions using templates describing movement with respect to the trajectory and objects. See the Appendix for details. Note that \crafty has an advantage over the learned models which rely on panoramic images to identify visual references and do not exploit object annotations.

\begin{table*}
\centering
\small
\setlength\tabcolsep{3.5pt}
\begin{tabularx}{\linewidth}{Xcccccccccc}
&&&&&&& & \multicolumn{2}{c}{\textbf{Visual Search}} \\
 \cmidrule{9-10}
\textbf{Instructions}  &   \textbf{Num. Evals} & \textbf{Wordcount} & \textbf{NE} $\downarrow$ & \textbf{SR} $\uparrow$ & \textbf{SPL} $\uparrow$ & \textbf{SDTW} $\uparrow$ & \textbf{Quality} $\uparrow$   &  \textbf{Start} $\downarrow$  &  \textbf{Other} $\downarrow$ & \textbf{Time} $\downarrow$ \\
\midrule  
\textbf{Val-unseen}  & & & & & \\
Speaker-Follower & 783$\times$3  &  24.6 & 6.55 & 35.8 & 30.3 & 28.1 & 3.50 & 43.5 & 24.8 & 43.2 \\
EnvDrop          & 783$\times$3  &  21.3 & 5.89 & 42.3 & 36.1 & 33.5 & 3.70 & 42.1 & 24.8 & \textbf{39.5} \\
\midrule
\textbf{Val-seen}   & & & & & \\
Speaker-Follower & 340$\times$3  &  24.7 & 6.23 & 42.3 & 35.7 & 33.2 & 3.64 & 43.0 & 24.5 & 42.6 \\
EnvDrop          & 340$\times$3  &  22.3 & 5.99 & 47.7 & 40.0 & 36.9 & 3.83 & 39.9 & 25.0 & 42.1 \\
Crafty       	 & 340$\times$3  &  71.2 & 6.01 & 43.6 & 34.7 & 33.3 & 3.48 & 42.0 & 25.9 & 69.6 \\
Direction Swap   & 340$\times$3  &  54.8 & 4.74 & 58.9 & 47.9 & 45.9 & 3.67 & 40.6 & 25.4 & 61.0 \\
Entity Swap      & 340$\times$3  &  55.1 & 4.71 & 51.3 & 42.5 & 40.6 & 3.33 & 40.1 & 25.8 & 62.7 \\
Phrase Swap      & 340$\times$3  &  52.0 & 4.07 & 62.6 & 51.6 & 49.9 & 3.85 & 38.7 & 24.7 & 58.0 \\
Human            & 340$\times$3  &  54.1 & \textbf{2.56} & \textbf{75.1} & \textbf{64.7} & \textbf{63.1} & \textbf{4.25} & \textbf{35.8} & \textbf{23.8} & 53.9 \\
\end{tabularx}
\caption{Human wayfinding performance following instructions from the Speaker-Follower \cite{fried2018speaker} and EnvDrop \cite{backtranslate2019} models, compared to Crafty (template-based) instructions, Human instructions, and three adversarial perturbations of Human instructions (Direction, Entity and Phrase Swap).}
\label{tab:human-eval}
\end{table*}

\paragraph{Human Instructions}
We collect 340 new English instructions for the trajectories in the R2R val-seen split using the PanGEA Guide task. 

\paragraph{Instruction Perturbations}

To quantify the impact of common instruction generator failure modes on instruction following performance, we include three adversarial perturbations of human instructions capturing incorrect direction words, hallucinated objects/landmarks, and repeated or skipped steps. We use Google Cloud NLP\footnote{\href{https://cloud.google.com/natural-language}{https://cloud.google.com/natural-language/}} to identify named entities and parse dependency trees and then generate perturbations as follows:

\begin{itemize}
\item \textbf{Direction Swap}: Random swapping of directional phrases with alternatives from the same set, with sets as follows: \textit{around/left/right, bottom/middle/top, up/down, front/back, above/under, enter/exit, backward/forward, away from/towards, into/out of, inside/outside}. 

Example: \textit{``Take a right (\textbf{left}) and wait by the couch outside (\textbf{inside}) the bedroom. "}

\item \textbf{Entity Swap}: Random swapping of entities in an instruction. All noun phrases excluding a stop list containing \textit{any, first, end, front}, etc. are considered to be entities. If two entities have the same lemma (e.g., \textit{stairs/staircase/stairway}) they are considered to be synonyms and are not swapped. 

Example: \textit{``Exit the bedroom (\textbf{bathroom}), turn right, then enter the bathroom (\textbf{bedroom})."}

\item \textbf{Phrase Swap}: 
A random operation on the dependency tree: either remove one sub-sentence tree, duplicate one sub-sentence tree, or shuffle the order of all sentences except the last.

Example: \textit{``Exit the room using the door on the left. Turn slightly left and go past the round table an chairs. Wait there."} -- where the first and second sentences are swapped.
\end{itemize}

\paragraph{Wayfinding Task}
Using the PanGEA Follower task, annotators are presented with a textual navigation instruction and the first-person camera view from the starting pose. They are instructed to attempt to follow the instruction to reach the goal location. Camera controls allow for continuous heading and elevation changes as well as movement between Matterport3D panoramas based on a navigation graph. 
Each instruction is evaluated by three different human wayfinders.

\paragraph{Evaluation Metrics}

We use the following standard metrics to evaluate the trajectories generated by our annotators (and hence, the quality of the provided instructions): 
\textit{Navigation Error} (\textbf{NE~$\downarrow$})
\textit{Success Rate} (\textbf{SR~$\uparrow$}),
\textit{Success weighted by inverse Path Length} (\textbf{SPL~$\uparrow$}),
\textit{Success weighted by normalized Dynamic Time Warping} (\textbf{SDTW~$\uparrow$}). Arrows indicate improving performance. See \citet{evaluation2018} and \citet{magalhaes2019effective} for details. 

People are resourceful and may succeed in following poor quality instructions by expending additional effort. Therefore, we report additional metrics to capture these costs. \textbf{Quality~$\uparrow$} is a self-reported measure of instruction quality based on a 1--5 Likert scale. At the end of each task annotators respond to the prompt: \textit{Do you think there are mistakes in the instruction?} Responses range from \textit{Way too many mistakes to follow (1)} to \textit{No mistakes, very very easy to follow' (5)}. \textbf{Visual Search} cost{~$\downarrow$} measures the percentage of the available panoramic visual field that the annotator observes at each viewpoint, based on the pose traces provided by PanGEA and first proposed in the RxR dataset \cite{ku-etal-2020-room}. Higher values indicate greater effort spent looking for the correct path. We report this separately for the start viewpoint and other viewpoints since wayfinders typically look around to orient themselves at the start. \textbf{Time~$\downarrow$} represents the average time taken in seconds.


\paragraph{Results}

Table \ref{tab:human-eval} summarizes the results of 11,886 wayfinding attempts using 37 English-speaking annotators. 
The performance of annotators stays consistent over time and does not show any sign of adaptation. See Appendix for detailed analysis.

As expected, human instructions perform best in human wayfinding evaluations on all path evaluation metrics and on subjective assessments of instruction quality, and they also incur the lowest visual search costs. The only metric not dominated by human instructions is the time taken -- which correlates with instruction length, and may be affected by wayfinders giving up when faced with poor quality instructions. Overall, the Speaker-Follower and EnvDrop models are surprisingly weak and noticeably worse than even adversarially perturbed human instructions. Compared to the template-based approach (Crafty), the Speaker-Follower model performs on par and EnvDrop is only slightly better.  As a first step to improving existing navigation instruction generators, we focus on developing and evaluating automated metrics that can approximate these human wayfinding evaluations.

\section{Compatibility Model}
\label{sec:model}

As an alternative to human evaluations, we train an instruction-trajectory compatibility model to assess both the grounding between textual and visual inputs and the alignment of the two sequences.

\subsection{Model Structure}

Our model is a dual encoder that encodes instructions and trajectories into a shared latent space (Figure \ref{fig:architecture}). The instruction representation $h^w$ is the concatenation of the final output states of a bi-directional LSTM \citep{Schuster1997BidirectionalRN} encoding the instruction tokens $\mathcal{W} = \{w_1, w_2, ..., w_n\}$. We use contextualized token embeddings from BERT \citep{devlin2018bert} as input to the LSTM.  

The visual encoder is a two-layer LSTM that processes visual features extracted from a sequence of viewpoints $\mathcal{V} = \{(I_1,p_1), (I_2,p_2) ..., (I_t,p_t)\}$ comprised of panoramic images $I_t$ captured at positions $p_t$ along a 3D trajectory. The vector $h^v_t$ representing the viewpoint at step $t$ is given by: 
\begin{align}
    a_t = \text{Attention}(h^v_{t-1}, e_{\text{pano}, t}) \\
    v_t = f([e_{\text{prev}, t}, e_{\text{next}, t}, a_t]) \\
    h^v_t = \text{LSTM}(v_t, h^v_{t-1})
\end{align}
\noindent
where $e_{\text{pano}, t}$ is a set of 36 visual features representing the panoramic image $I_t$ (discretized into 36 viewing angles by elevation $\theta$ and heading $\phi$), $e_{\text{prev}, t}$ and $e_{\text{next}, t}$ are the visual features in the directions of the previous and next viewpoints ($v_{t-1}-v_t$ and $v_{t+1}-v_t$ respectively), and $f$ is a projection layer. Each visual feature is a concatenation of a pre-trained CNN image feature \cite{starburst} with orientation vectors encoding both sine and cosine functions of the absolute and relative angles $\{ \theta_\text{abs}, \phi_\text{abs}, \theta_\text{rel}, \phi_\text{rel} \}$. 
We use standard dot-product attention \citep{luong2015effective} and define $h^v = h^v_T$, the final viewpoint embedding in the trajectory.
The output of the model is the compatibility score $S$ between an instruction and a trajectory defined as the cosine similarity between $h^v$ and $h^w$.

\subsection{Hard Negative Mining}
To avoid overfitting, our compatibility model is \textit{not trained on the outputs of any of the instruction generators that we evaluate}. Instead, we use only the relatively small set of positive instruction-trajectory examples from R2R. We use round-trip back-translation to expand the set of positive examples. Unmatched instruction-trajectory pairs from R2R are considered to be negative examples and we also construct hard negative examples from positive examples by adversarially perturbing both trajectories and instructions. 

\paragraph{Instruction Perturbations}
We use the same instruction perturbations described in Section \ref{sec:human_eval}: \textit{Direction Swap}, \textit{Entity Swap}, and \textit{Phrase Swap}. These perturbations are inspired by typical failure modes in instruction generators and are designed to be hard to recognize without grounding on images and actions along the trajectory. Previous work by \citet{Huang2019Transferable} considered only trajectory perturbations. While this encourages the model to recognize incorrect trajectories for a given ground truth instruction, it may not encourage the model to identify a trajectory matched with a poor quality instruction. Our results suggest that instruction perturbations are equally important. 

\paragraph{Trajectory Perturbations} 
To perturb trajectories we use the navigation graphs defining connected viewpoints in R2R. Inspired by \citet{Huang2019Transferable}, we consider \textit{Random Walk}, \textit{Path Reversal}, and  \textit{Viewpoint Swap} perturbations:
\begin{itemize}
\item \textbf{Random Walk}: The first or last two viewpoints are fixed and the remainder of the trajectory is re-sampled using random edge traversals subject to the path length remaining within $\pm1$ step of the original. To make the task harder, we avoid revisiting a viewpoint and require the re-sampled trajectory to have at least two overlapping viewpoints with the original.
\item \textbf{Path Reversal}: The entire trajectory is reversed while keeping the same viewpoints.
\item \textbf{Viewpoint Swap}: A new method we introduce that randomly samples and swaps a viewpoint in a trajectory with a new viewpoint sampled from the neighbors of the adjacent viewpoints in the original trajectory. 
\end{itemize}

\paragraph{Paraphrases}
To expand the 14k positive examples from the R2R train set and balance the positive-to-negative ratio, we paraphrase instructions via round-trip back-translation. We use the following ten intermediate languages and Google Translate\footnote{\href{https://cloud.google.com/translate/}{https://cloud.google.com/translate/}}: \textit{ar, es, de, fr, hi, it, pt, ru, tr}, and \textit{zh}. To exclude low quality or nearly duplicate instructions, we filter paraphrased instructions outside the BLEU score range of [0.25, 0.7] compared to the original. Overall we have a total of 110,601 positive instruction-trajectory pairs in the training set, which contains 4675 unique trajectories.

\begin{table*}
\centering
\scriptsize
\setlength\tabcolsep{2pt}
\begin{tabularx}{\linewidth}{lXccccccccc}
&&& & & \multicolumn{3}{c}{\textbf{Instruction Validation}} & \multicolumn{3}{c}{\textbf{Path Validation}} \\
\cmidrule(l){6-8} 
\cmidrule(l){9-11} 
 & \textbf{Model} & \textbf{Perturbations}  & \textbf{Validation} & \textbf{Test} & Direction & Entity & Phrase & Viewpoint & Random & Path\\
 &                &                         & (val\_unseen)&(val\_seen)& Swap & Swap & Swap & Swap & Walk & Reversal \\
\midrule  
1 & \citet{Huang2019Transferable}  & Path + Instruction  & 53.4 & 52.3 & 	80.3 & 89.4 & 80.9 & 78.0 & 72.9 & 75.9  \\
\midrule
  &\textbf{This Work}   \\
2 & ~+ CE Loss     & Path + Instruction & 57.9 & 57.6 & 89.5 & 88.4 & 83.4	& 95.0 & 94.1 & 87.8 \\
3 & ~+ Focal Loss  & Path + Instruction & 59.2	& 59.2 & \textbf{89.8} &	90.5 & 84.2	& 95.6 & 95.0 & 90.3 \\
4 & ~+ Contrastive Loss  & Path + Instruction & 67.2 & 68.7 & 75.1	& 70.1 & 73.7 &	73.6 & 88.7 & 93.1 \\
5 & ~+ Contrastive + CE  & Path + Instruction & 66.5 & 67.5 & 82.0 & 77.4 & 76.9 & 84.7 & 91.3 & 91.7 \\
6 & ~+ Contrastive + Focal  & Path + Instruction  & 68.5 & 68.3 & 83.9 & 81.5 & 79.8 & 88.5 & 93.7 & 93.3 \\
7 & ~+ Contrastive + Focal + Paraphrase  & Path + Instruction  & 71.3 & 72.2 & 83.1 & 81.9 & 80.4 & 90.4 & 95.2 & \textbf{94.0} \\ %
8 & ~+ Contrastive + Focal + Paraphrase + Bert Embed.  & Path + Instruction  & \textbf{73.5} & \textbf{73.7} & 84.6 &	86.7 & 82.2 & 89.1 & 94.3 & 93.3 \\
\midrule  
& ~\textbf{Perturbation Ablations}   \\
9 & ~+ Contrastive + Focal + Paraphrase + Bert Embed.  & Instruction           & 70.5 & 70.4 & 85.1 & 88.6 & 82.5 & 64.9 & 84.3 & 90.7 \\ 
10 & ~+ Contrastive + Focal + Paraphrase + Bert Embed.  & Direction Swap Only   & 70.1 & 68.9 & 89.0 & 72.0 & 70.7 &	65.9 & 85.0	& 91.0 \\
11 & ~+ Contrastive + Focal + Paraphrase + Bert Embed.  & Entity Swap Only      & 69.1 &	68.5 & 70.3 & \textbf{92.7} & 71.3 & 65.2 & 84.4 & 90.9 \\
12 & ~+ Contrastive + Focal + Paraphrase + Bert Embed.  & Phrase Swap Only      & 70.3 &	70.5 & 70.1	& 72.5 & \textbf{85.5} & 65.5 & 83.3 & 90.1\\
13 & ~+ Contrastive + Focal + Paraphrase + Bert Embed.  & Path   & 69.7 & 69.4  & 71.7 & 71.4 & 69.8 & 92.4 & 94.8 & 92.2 \\
14 & ~+ Contrastive + Focal + Paraphrase + Bert Embed.  & Viewpoint Swap Only   & 70.7 & 72.1 & 70.7 & 71.1 & 71.2 & \textbf{94.6} & 94.6 & 91.9 \\
15 & ~+ Contrastive + Focal + Paraphrase + Bert Embed.  & Random Walk Only      & 70.5 & 70.9 & 70.5 & 71.5 & 71.0 & 81.7 & \textbf{95.1} & 91.6 \\
16 & ~+ Contrastive + Focal + Paraphrase + Bert Embed.  & Path Reversal Only    & 69.4 & 69.7 & 71.0 & 70.7 & 71.2 & 65.3 & 86.3 & 91.9 \\
17 & ~+ Contrastive + Focal + Paraphrase + Bert Embed.  & No Perturbation                  & 69.1 & 69.7 &	71.1 & 71.0 & 70.8 & 65.3 & 84.9 & 90.2 \\
\end{tabularx}
\caption{Ablation of different models based on classification AUC. Models are trained with the original R2R-train data  and paraphrased positive instructions, plus path and/or instruction perturbed hard negatives (second column). The best models are selected based on the validation set (column 3), and we report the final test performance in column 4. To understand the performance of individual perturbation method, we also report the best AUCs for each of the six perturbations in columns 5 - 10.}
\label{tab:model-scores}
\end{table*}

\subsection{Loss Functions}

During training, each minibatch is constructed with $N$ matching instruction-trajectory pairs, which may be perturbed. We define $M \in {\{ 0, 1 \}}^{N}$ as the vector indicating unperturbed pairs. A compatibility matrix $S \in \mathcal{R}^{N \times N}$ is defined such that $S_{i,j}$ is the cosine similarity score between instruction $i$ and trajectory $j$ determined by our model.  

We use both binary classification loss functions, defined on diagonal elements of $S$, and a contrastive loss defined on $S$'s rows and columns. Contrastive losses are commonly used for retrieval and representation learning \citep[e.g.,][]{Yang2019multilingual, Chen2020SimCLR} and in our case exploits all random instruction-trajectory pairs in a minibatch. 

Each loss requires a separate normalization. For the classification loss we compute the probability of a match $p_{i,j}$, such that $p_{i,j} = \sigma(aS_{i,j} + b)$ where $a$ and $b$ are learned scalars and $\sigma$ is the sigmoid function. For the classification loss $\mathcal{L}_{\text{cls}}$ we consider both binary cross entropy loss $\mathcal{L}_{\text{CE}}$, and focal loss \citep{lin2018focal} given by $\mathcal{L}_{\text{FL}} = (1-p_{i,j})^\gamma\mathcal{L_{\text{CE}}}$ where we set $\gamma=2$. 

For the contrastive loss we compute logits by scaling $S$ with a learned scalar temperature $\tau$. The contrastive loss $\mathcal{L}_C(S)$ calculated over the rows and colums of $S$ is given by:
\begin{align}
    \mathcal{L}_C(S) = \frac{1}{\sum_i M_{i}} \sum_{i=1}^N \Big(\mathcal{L}_{r}(S_i) + \mathcal{L}_{r}(S^\intercal_i) \Big)
\end{align}
where $\mathcal{L}_{r}(S_i)=0$ if $M_{i} = 0$, i.e.,  the diagonal element is a perturbed pair and not considered to be a match. Otherwise:
\begin{align}
    \mathcal{L}_{r}(S_i) = -\log\frac{e^{S_{i,i}/\tau}}{\sum_j^N  e^{S_{i,j}/\tau}}
\end{align}
The final loss is the combination:
\begin{align}
    \mathcal{L} = \mathcal{L}_{C}(S) + \frac{\beta}{N} \sum_{i=1}^N\mathcal{L}_{cls}(S_{i,i})
\end{align}
where $\mathcal{L}_{cls}$ is the classification loss, either $\mathcal{L}_{CE}$ or $\mathcal{L}_{FL}$, and we set $\beta=1$.

\paragraph{Sampling hyperparameters}
 
We sample positive and negative examples equally with a mix ratio of 2:1:1 for ground truth, instruction perturbations, and trajectory perturbations, respectively. For each perturbation type, we sample the three methods with equal probability.

\section{Experiments}
\label{sec:exp}

We evaluate our compatibility model against alternative model-based evaluations and standard textual similarity metrics. We report instruction classification results in Section \ref{sec:classification}, improved data augmentation for VLN agents in Section \ref{sec:vln}, and correlation with human wayfinder outcomes in \ref{sec:metrics}.

\subsection{Instruction Classification}
\label{sec:classification}

\begin{figure*}
\centering
\scriptsize
\includegraphics[width=0.328\textwidth]{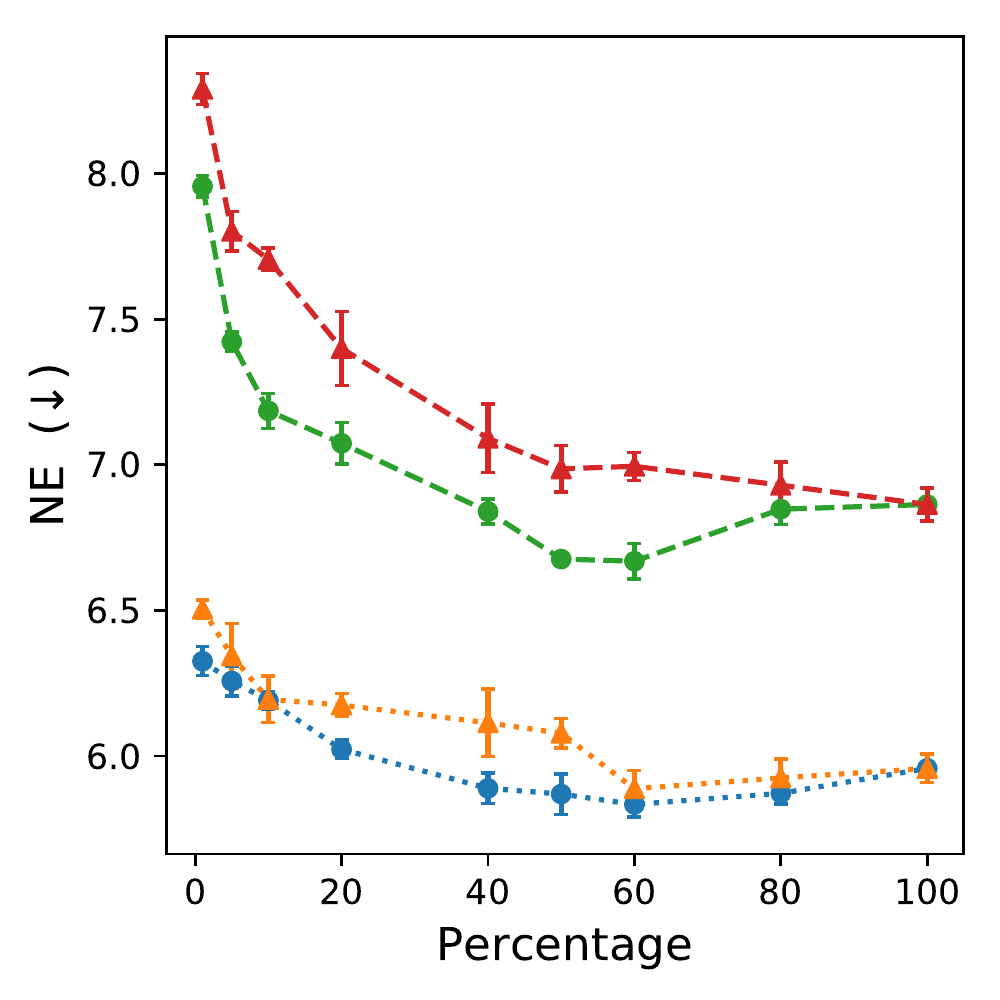}
\includegraphics[width=0.328\textwidth]{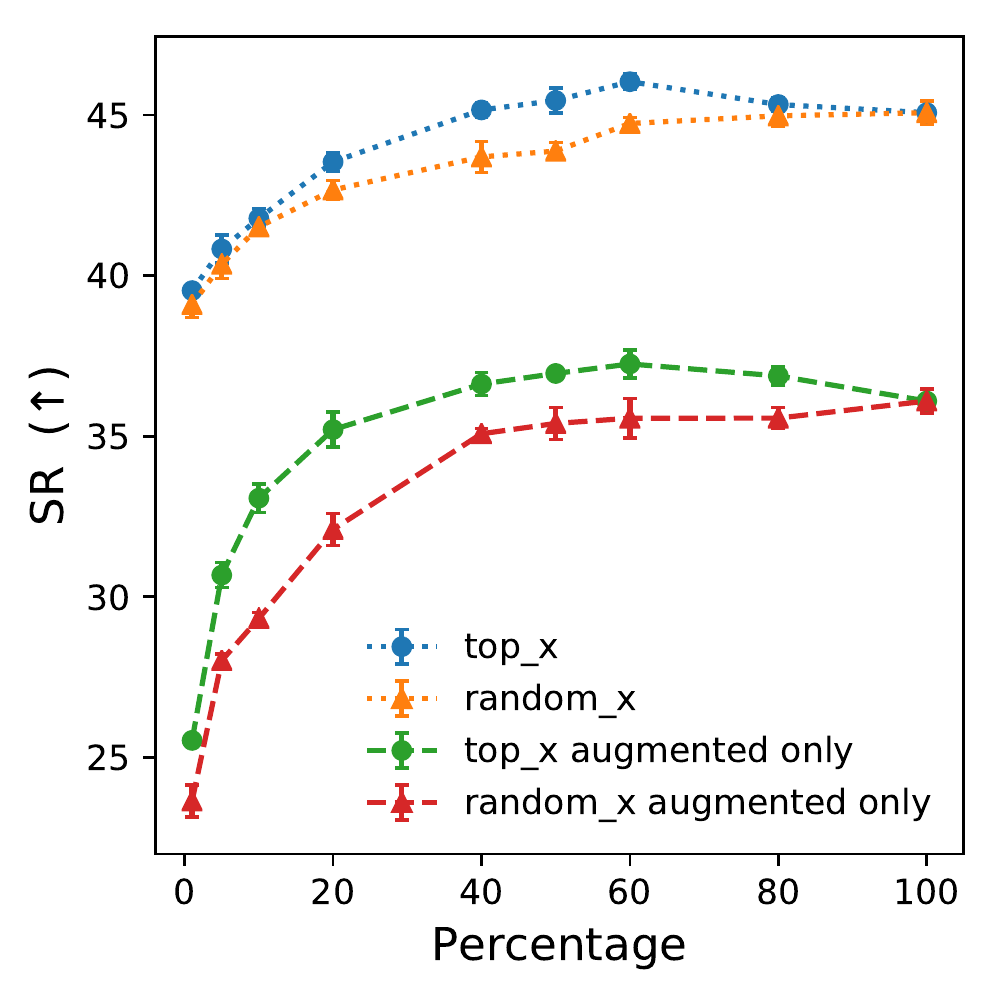}
\includegraphics[width=0.328\textwidth]{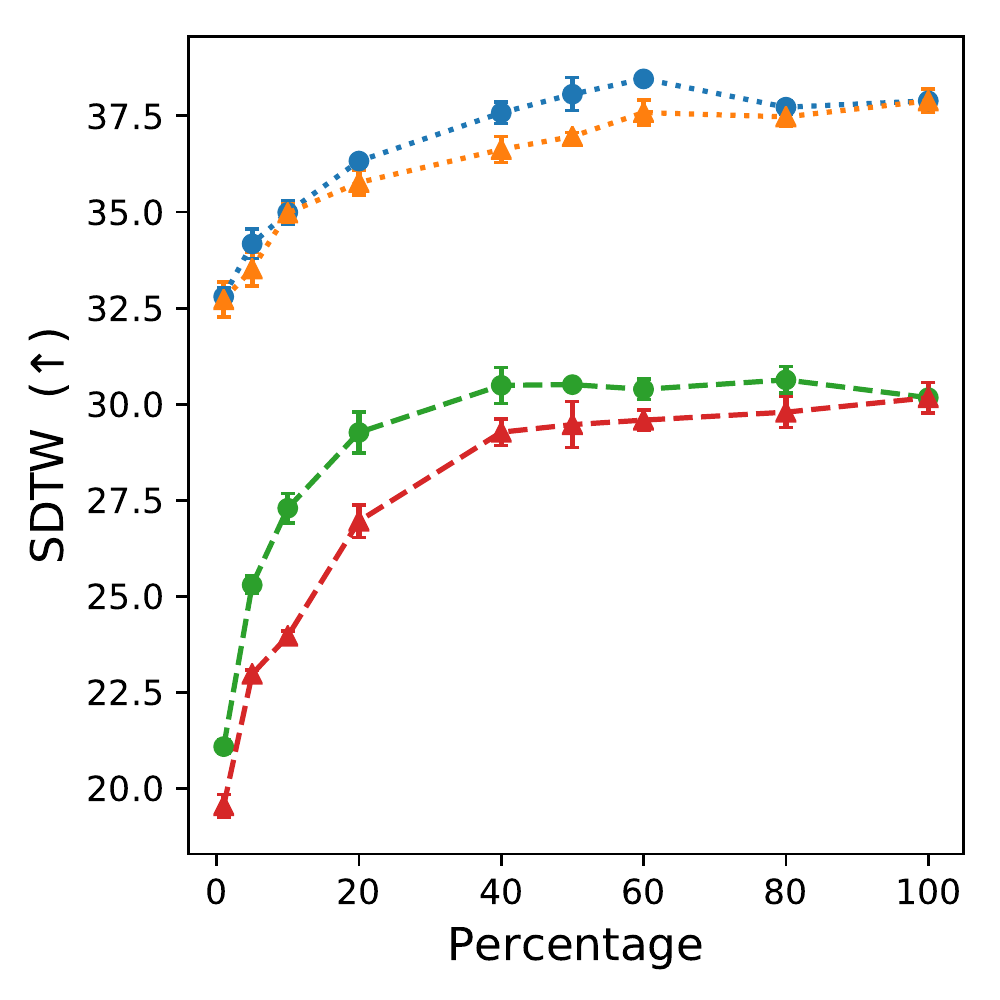}
\caption{Navigation performance of a VLN agent trained with different fractions of Speaker-Follower augmented paths, starting from 1\%. For NE, lower is better; for SR and SDTW, higher is better. The dashed-lines (green and red) use only augmented paths in training, while the dotted lines (blue and orange) use both augmented and R2R-train paths. Filled circles indicate fractions ranked by our compatibility model in descending order, while triangles indicate random fractions. Each point is the mean of 3 runs and the error bars represent the standard deviation of the mean. The model-ranked fractions show consistent improvement over random samples of the same percentage. Agents trained with only augmented paths (dashed-lines) show greater difference between model-ranked fractions and random fractions.}
\label{fig:vln_perf}
\end{figure*}

\paragraph{Evaluation}
In this setting we use the instruction-trajectory compatibility model to classify high and low quality instructions for trajectories from the R2R val-unseen and val-seen sets. The instruction pool includes 3 high-quality instructions per trajectory from R2R, plus 2 instructions per trajectory from the Speaker-Follower and EnvDrop models. These are considered to be high quality if 2 out of 3 human wayfinders reached the goal (see Section \ref{sec:human_eval}), and low quality otherwise. We assess model performance using \textit{Area Under the ROC Curve} (AUC). We use the val-unseen split (3915 instructions, 75\% high quality) for model validation and the val-seen split (1700 instructions, 78\% high quality) as test.

\paragraph{Benchmark}
We compare to the compatibility model proposed by \citet{Huang2019Transferable}, which computes elementwise similarities between instruction words and trajectory panoramas \textit{before} pooling, and does not include action embeddings ($e_{\text{prev}, t}$ and $e_{\text{next}, t}$) or position encodings $p_t$. In contrast, our model calculates the similarity between instructions and trajectories \textit{after} pooling each sequence and includes both action and position encodings.

\paragraph{Results}
Table \ref{tab:model-scores} reports classification AUC including comprehensive ablations of loss functions, approaches to hard negative mining, and modeling choices. With regard to the loss function, we find that the combination of contrastive and focal loss (row 6) performs best overall, and that adding contrastive loss provides a very significant $9$ - $10\%$ increase in AUC compared to using just cross-entropy (CE) or focal loss (rows 2 and 3) due to the effective use of in-batch negatives. Adding paraphrased positive instructions and pretrained BERT token embeddings also leads to significant performance gains (rows 7 and 8 vs. row 6). The best performing model on both the validation and test sets uses Contrastive + Focal loss with paraphrased instructions and BERT embeddings, as well as trajectory and instruction perturbations (row 8). This model consistently outperforms the benchmark from prior work (row 1) by a large margin and achieves a test set AUC of 73.7\%. 

In rows 9--17 we ablate the six perturbation methods that we use for hard negative mining. Ablations using only instruction perturbations (row 9), only path perturbations (13), or no perturbations at all (row 17) perform considerably worse than our best model (row 8). We also show that no individual perturbation approach is effective on its own. In addition to scores for the validation and test sets, we report AUC for each perturbation method on the val-seen set to investigate their individual performance. Overall, trajectory perturbations get higher scores than instruction perturbations, showing they are easier tasks. \textit{Phrase Swap} proves the hardest task, while \textit{Random Walk} is the easiest.

\subsection{Data Augmentation for VLN}
\label{sec:vln}

Data augmentation using instructions from the Speaker-Follower and EnvDrop models is pervasive in the training of VLN agents \cite{wang2018reinforced,Huang2019Transferable,li2019robust}. In this section we evaluate whether our compatibility model can be used to filter out low quality instructions from the augmented training set to improve VLN performance. 
We score all of 170k augmented instruction-trajectory pairs from the Speaker-Follower model and rank them in descending order. We then use different fractions of the ranked data to train VLN agents, and compare with agents trained using random samples of the same size. We use a VLN agent model based on \citet{wang2018reinforced} and implemented in VALAN \cite{lansing2019valan}, which achieves a success rate (SR) of ~45\% on the R2R val-unseen split when trained on the R2R train split and all of the Speaker-Follower augmented instructions.

Figure \ref{fig:vln_perf} indicates that instruction-trajectory pairs selected by our compatibility model consistently outperform random training pairs in terms of the performance of the trained VLN agent. 
This demonstrates the efficacy of our compatibility model for improving VLN data augmentation by identifying high quality instructions.

\subsection{Correlation with Human Wayfinders}
\label{sec:metrics}

In this section we evaluate the correlation between the scores given by our instruction-trajectory compatability model and the outcomes from the human wayfinding attempts described in Section \ref{sec:human_eval}. Using Kendall's $\tau$ to assess rank correlation, we report both system-level and instance-level correlation. The instance-level evaluations assess whether the metric can identify the best \textit{instruction} from two candidates, while the system-level evaluations assess whether a metric can identify the best \textit{model} from two candidates (after averaging over many instruction scores for each model). The results in Table \ref{tab:corr} are reported separately over all 3.9k instructions (9 systems comprising the rows of Table \ref{tab:human-eval}), and over model-generated instructions only (4 systems comprising the 2.2k instructions generated by the Speaker-Follower and EnvDrop models on R2R val-seen and val-unseen). 

\paragraph{Automatic Metrics}
For comparison we include standard textual evaluation metrics (BLEU, CIDEr, METEOR, ROUGE and SPICE) and two model-based metrics: BERTScore \citep{Zhang2019bertscore}, and scores based on the performance of a trained VLN agent attempting to follow the candidate instruction \cite{vls-2019}. Note that only the compatibility model and the VLN agent-based scores use the candidate trajectory -- the other metrics are calculated by comparing each candidate instruction to the three reference instructions from R2R (and are thus reliant on reference instructions). 

To calculate the standard metrics we use the official evaluation code provided with the COCO captions dataset \cite{Chen2015}. For BERTScore, we use a publicly available uncased BERT\footnote{\href{https://tfhub.dev/google/bert_uncased_L-12_H-768_A-12/1}{tfhub.dev/google/bert\_uncased\_L-12\_H-768\_A-12/1}} model with 12 layers and hidden dimension 768, and compute the mean $F1$-score over the three references.  
For the VLN agent score, we train three VLN agents based on \citet{wang2018reinforced} from different random initializations using the R2R train set. We then employ the trained agents for the wayfinding task and report performance as either the SPL or SDTW similarity between the path taken by the agent and the reference path -- using either a single agent or the average score from three agents.

\begin{table}[t]
\centering
\scriptsize
\setlength\tabcolsep{1pt}
\begin{tabularx}{\linewidth}{lXcrlcrlcrlcrl}
&&& \multicolumn{11}{c}{\textbf{All Instructions (N=3.9k, M=9)}}\\
\cmidrule{4-14}
& \textbf{Score} & \textbf{Ref} & \multicolumn{2}{c}{\textbf{NE} $\downarrow$} & & \multicolumn{2}{c}{\textbf{SR} $\uparrow$} & & \multicolumn{2}{c}{\textbf{SPL} $\uparrow$} & & \multicolumn{2}{c}{\textbf{Quality} $\uparrow$}\\
\midrule  
\multirow{9}{*}{\rotatebox[origin=c]{90}{\textbf{System-Level}}}
& BLEU-4      & \checkmark & (\pn0.00, & \pn0.33) & & (-0.22, & \pn0.39) & & (-0.22, & \pn0.00) & & (\pn0.11, & \pn0.39)\\
& CIDEr       & \checkmark & (\pn0.06, & \pn0.39) & & (-0.22, & \pn0.39) & & (-0.22, & \pn0.00) & & (\pn0.17, & \pn0.39)\\
& METEOR      & \checkmark & (\pn0.11, & \pn0.44) & & (-0.39, & \pn0.28) & & (-0.39, & -0.06) & & (\pn0.00, & \pn0.28)\\
& ROUGE       & \checkmark & (\pn0.06, & \pn0.39) & & (-0.28, & \pn0.39) & & (-0.33, & \pn0.00) & & (\pn0.06, & \pn0.39)\\
& SPICE       & \checkmark & (\textbf{-0.67}, & \textbf{-0.28}) & & (\textbf{-0.06}, & \pn\textbf{0.61}) & & (\pn\textbf{0.44}, & \pn\textbf{0.78}) & & (\pn\textbf{0.56}, & \pn\textbf{0.83})\\
& BERTScore   & \checkmark & (\pn0.06, & \pn0.39) & & (-0.22, & \pn0.39) & & (-0.22, & \pn0.00) & & (\pn0.17, & \pn0.39)\\
& SPL$_{\text{1-agent}}$     & & (-0.50, & -0.06) & & (-0.22, & \pn0.44) & & (\pn0.11, & \pn0.56) & & (\pn0.00, & \pn0.44)\\
& SPL$_{\text{3-agents}}$      & & (-0.22, & \pn0.17) & & (-0.33, & \pn0.39) & & (\pn0.00, & \pn0.33) & & (\pn0.33, & \pn0.61)\\
& SDTW$_{\text{1-agent}}$      & & (-0.44, & \pn0.00) & & (-0.22, & \pn0.44) & & (\pn0.11, & \pn0.50) & & (\pn0.00, & \pn0.44)\\
& SDTW$_{\text{3-agents}}$      & & (-0.22, & \pn0.17) & & (-0.28, & \pn0.33) & & (\pn0.00, & \pn0.33) & & (\pn0.33, & \pn0.61)\\
& Compatibility          & & (-0.17, & \pn0.17) & & (-0.17, & \pn0.50) & & (\pn0.00, & \pn0.28) & & (\pn0.44, & \pn0.72)\\
\midrule
&&& \multicolumn{11}{c}{\textbf{All Instructions (N=3.9k, M=9)}}\\
\cmidrule{4-14}
& \textbf{Score} & \textbf{Ref} & \multicolumn{2}{c}{\textbf{NE} $\downarrow$} & & \multicolumn{2}{c}{\textbf{SR} $\uparrow$} & & \multicolumn{2}{c}{\textbf{SPL} $\uparrow$} & & \multicolumn{2}{c}{\textbf{Quality} $\uparrow$}\\
\midrule  
\multirow{9}{*}{\rotatebox[origin=c]{90}{\textbf{Instance-Level}}}
& BLEU-4      & \checkmark & (\pn0.05, & \pn0.09) & & (-0.04, & \pn0.00) & & (-0.09, & -0.05) & & (-0.01, & \pn0.03)\\
& CIDEr       & \checkmark & (\pn0.06, & \pn0.09) & & (-0.04, & -0.00) & & (-0.11, & -0.07) & & (-0.02, & \pn0.01)\\
& METEOR      & \checkmark & (\pn0.00, & \pn0.04) & & (-0.05, & -0.02) & & (-0.04, & \pn0.00) & & (-0.01, & \pn0.02)\\
& ROUGE     & \checkmark & (\pn0.05, & \pn0.08) & & (-0.05, & -0.01) & & (-0.10, & -0.06) & & (-0.02, & \pn0.02)\\
& SPICE       & \checkmark & (-0.05, & -0.02) & & (-0.00, & \pn0.04) & & (\pn0.03, & \pn0.06) & & (\pn0.03, & \pn0.07)\\
& BERTScore   & \checkmark & (-0.04, & -0.00) & & (\pn0.07, & \pn0.12) & & (-0.01, & \pn0.03) & & (\pn0.07, & \pn0.11)\\
& SPL$_{\text{1-agent}}$     & & (-0.18, & -0.14) & & (\pn0.15, & \pn0.19) & & (\pn0.14, & \pn0.18) & & (\pn0.07, & \pn0.11)\\
& SPL$_{\text{3-agents}}$      & & (\textbf{-0.22}, & \textbf{-0.18}) & & (\pn\textbf{0.20}, & \pn\textbf{0.24}) & & (\pn\textbf{0.18}, & \pn\textbf{0.22}) & & (\pn0.10, & \pn0.14)\\
& SDTW$_{\text{1-agent}}$     & & (-0.18, & -0.14) & & (\pn0.15, & \pn0.19) & & (\pn0.14, & \pn0.18) & & (\pn0.08, & \pn0.12)\\
& SDTW$_{\text{3-agents}}$     & & (\textbf{-0.22}, & \textbf{-0.19}) & & (\pn\textbf{0.20}, & \pn\textbf{0.24}) & & (\pn\textbf{0.18}, & \pn\textbf{0.22}) & & (\pn0.11, & \pn0.15)\\
& Compatibility          & & (\textbf{-0.20}, & \textbf{-0.17}) & & (\pn0.13, & \pn0.17) & & (\pn\textbf{0.17}, & \pn\textbf{0.20}) & & (\pn\textbf{0.19}, & \pn\textbf{0.23})\\
\midrule
&&& \multicolumn{11}{c}{\textbf{Model-Generated Instructions (N=2.2k, M=4)}}\\
\cmidrule{4-14}
& \textbf{Score} & \textbf{Ref} & \multicolumn{2}{c}{\textbf{NE} $\downarrow$} & & \multicolumn{2}{c}{\textbf{SR} $\uparrow$} & & \multicolumn{2}{c}{\textbf{SPL} $\uparrow$} & & \multicolumn{2}{c}{\textbf{Quality} $\uparrow$}\\
\midrule  
\multirow{9}{*}{\rotatebox[origin=c]{90}{\textbf{Instance-Level}}}
& BLEU-4      & \checkmark & (-0.02, & \pn0.03) & & (-0.03, & \pn0.02) & & (-0.02, & \pn0.03) & & (-0.02, & \pn0.03) \\
& CIDEr       & \checkmark & (-0.02, & \pn0.03) & & (-0.03, & \pn0.02) & & (-0.02, & \pn0.03) & & (-0.02, & \pn0.03) \\
& METEOR      & \checkmark & (-0.02, & \pn0.03) & & (-0.03, & \pn0.02) & & (-0.02, & \pn0.03) & & (-0.02, & \pn0.03) \\
& ROUGE       & \checkmark & (-0.02, & \pn0.03) & & (-0.05, & \pn0.00) & & (-0.04, & \pn0.01) & & (-0.03, & \pn0.02)\\
& SPICE       & \checkmark & (-0.05, & -0.00) & & (\pn0.00, & \pn0.05) & & (\pn0.00, & \pn0.05) & & (\pn0.01, & \pn0.06) \\
& BERTScore   & \checkmark & (-0.22, & -0.18) & & (\pn0.19, & \pn0.24) & & (\pn0.18, & \pn0.23) & & (\pn0.16, & \pn0.20) \\
& SPL$_{\text{1-agent}}$      & & (-0.21, & -0.16) & & (\pn0.17, & \pn0.23) & & (\pn0.16, & \pn0.22) & & (\pn0.07, & \pn0.12) \\
& SPL$_{\text{3-agents}}$      & & (\textbf{-0.26}, & \textbf{-0.21}) & & (\pn\textbf{0.21}, & \pn\textbf{0.27}) & & (\pn\textbf{0.21}, & \pn\textbf{0.26}) & & (\pn0.09, & \pn0.14)  \\
& SDTW$_{\text{1-agent}}$      & & (-0.22, & -0.16) & & (\pn0.17, & \pn0.23) & & (\pn0.16, & \pn0.22) & & (\pn0.07, & \pn0.13) \\
& SDTW$_{\text{3-agents}}$      & & (\textbf{-0.26}, & \textbf{-0.21}) & & (\pn\textbf{0.22}, & \pn\textbf{0.27}) & & (\pn\textbf{0.21}, & \pn\textbf{0.26}) & & (\pn0.10, & \pn0.15) \\
& Compatibility          & & (\textbf{-0.25}, & \textbf{-0.20}) & & (\pn\textbf{0.22}, & \pn\textbf{0.27}) & & (\pn\textbf{0.21}, & \pn\textbf{0.25}) & & (\pn\textbf{0.18}, & \pn\textbf{0.23}) \\
\end{tabularx}
\caption{Kendall's $\tau$ correlation between automated instruction evaluation metrics and human wayfinder evaluations. Ranges are 90\% confidence intervals based on bootstrap resampling. \textbf{N} refers to the number of instructions and \textbf{M} refers to the number of systems. If checked, \textbf{Ref} indicates that the metric requires reference instructions for comparison. SPL\textsubscript{$k$-agent(s)} and SDTW\textsubscript{$k$-agent(s)} refers to wayfinding scores averaged over $k$ VLN agents trained from random initialization. 
}
\label{tab:corr}
\end{table}

\paragraph{Results} 

\begin{figure}
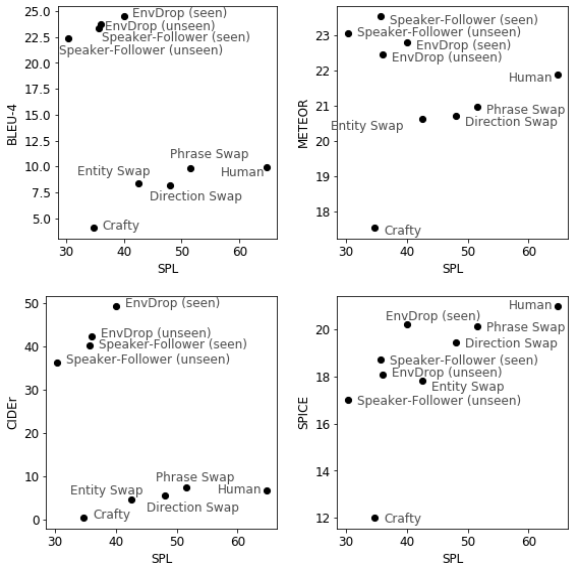

\centering
\scriptsize
\includegraphics[width=0.49\columnwidth]{figs/bleu-4.pdf}
\includegraphics[width=0.49\columnwidth]{figs/meteor.pdf}
\includegraphics[width=0.49\columnwidth]{figs/cider.pdf}
\includegraphics[width=0.49\columnwidth]{figs/spice.pdf}
\caption{Standard evaluation metrics vs. human wayfinding outcomes (SPL) for 9 navigation instruction generation systems. SPICE is most consistent with human wayfinding outcomes, although no metrics score the Crafty template-based instructions highly.}
\label{fig:scatterplots}
\end{figure}

Table \ref{tab:corr} compares system-level and instance-level correlations for all metrics, both standard and model-based. 
At the \textit{system-level}, we see no correlation between standard text metrics such as BLEU, ROUGE, METEOR and CIDEr and human wayfinder performance. The exception is SPICE, which shows the desired negative correlation with NE, and positive correlation with SR, SPL (see Figure \ref{fig:scatterplots}) and Quality. At the system-level, the model-based approaches (BERTScore, agent SPL/SDTW and Compatability) also lack the desired correlation and exhibit wide confidence intervals. Here, it is important to point out that the 9 systems under evaluation include a variety of styles (e.g., Crafty's template-based instructions, different annotator pools, adversarial perturbations) which are dissimilar to the R2R data used to train the VLN agents and the compatibility model. Accordingly, the model-based approaches are unable to reliably rank these out-of-domain systems.

At the \textit{instance-level} (when scoring individual instructions) we observe different outcomes. SPICE scores for individual instructions have high variance, and so SPICE does not correlate with wayfinder performance at the instruction level. In contrast, the model-based approaches exhibit the desired correlation, particularly when restricted to the model-generated instructions (Table \ref{tab:corr} bottom panel). Our compatibility score shows the strongest correlation among all metrics, performing similarly to an ensemble of three VLN agents.

\section{Conclusion}
\label{sec:conclusion}

Generating grounded navigation instructions is one of the most promising directions for improving the performance of VLN wayfinding agents, and a challenging and important language generation task in its own right. In this paper, we show that efforts to improve navigation instruction generators have been hindered by a lack of suitable automatic evaluation metrics. With the exception of SPICE, all the standard textual evaluation metrics we evaluated (BLEU, CIDEr, METEOR and ROUGE) are ineffective, and -- perhaps as a result -- existing instruction generators have substantial headroom for improvement.

To address this problem, we develop an instruction-trajectory compatibility model that outperforms all existing automatic evaluation metrics on instance-level evaluation without needing any reference instructions -- making it suitable for use as a reward function in a reinforcement learning setting, as a discriminator in a Generative Adversarial Network (GAN) \cite{dai2017towards}, or for filtering instructions in a data augmentation setting. 

Progress in natural language generation (NLG) is increasing the demand for evaluation metrics that can accurately evaluate generated text in a variety of domains. Our findings are a timely reminder that textual evaluation metrics should not be trusted in new domains unless they have been comprehensively validated against human judgments. In the case of grounded navigation instructions, for model selection in the presence of reference instructions we recommend using the SPICE metric. In all other scenarios (e.g., selecting individual instructions, or model selection without reference instructions) we recommend using a learned instruction-trajectory compatibility model.

\section*{Acknowledgements}

We thank the Google Data Compute team, in particular Ashwin Kakarla and Priyanka Rachapally, for their tooling and annotation support for this project.

\bibliography{paper}
\bibliographystyle{acl_natbib}

\clearpage
\appendix

\section{Automated metric scores for all instructions}

We provide more details about automated metric scores for all instructions in this section.
Table \ref{tab:metric-scores} gives automated metrics for each model we consider. Generated instructions from EnvDrop and Speaker-Follower are scored the highest, whereas human instructions are scored poorly and on par with perturbed instructions, and Crafty is the lowest. These results diverge significantly from human wayfinding performance in Section 3, and highlights the inefficacy of these automated text metrics.

\begin{table}[h!]
\centering
\scriptsize
\setlength\tabcolsep{2pt}
\begin{tabularx}{\linewidth}{lcccccc}
\textbf{}  &    \textbf{BertScore}  &\textbf{BLEU-4} & \textbf{CIDEr} & \textbf{ROUGE} & \textbf{METEOR} & \textbf{SPICE}  \\
\midrule  
\textbf{Val-unseen}  \\
Speaker Fol.     & 78.9  &   22.3 & 36.3 & 45.6 & 23.1 & 17.0 \\
EnvDrop          & 79.3 & 23.7 & 42.2 & 45.8 & 22.5 & 18.1 \\
\midrule
\textbf{Val-seen}   & & & & & & \\
Speaker Fol.     & 79.0 & 23.3 & 40.1 & 46.3 & \textbf{23.5} & 18.7 \\
EnvDrop          & \textbf{79.5} & \textbf{24.5} & \textbf{49.3} & \textbf{46.7} & 22.8 & 20.2 \\
Crafty           & 71.5 & 4.1 & 0.4 & 23.3 & 17.5 & 12.0 \\
Dir. Swap        & 74.7 & 8.2 & 5.7 & 31.4 & 20.7 & 19.4 \\
Entity Swap      & 74.0 & 8.4 & 4.6 & 31.4 & 20.6 & 17.8 \\
Phrase Swap      & 74.8 & 9.8 & 7.4 & 31.4 & 21.0 & 20.1 \\
Human            & 74.9 & 9.9 & 6.7 & 33.3 & 21.9 & \textbf{21.0} \\
\end{tabularx}
\caption{Automated metric scores for all instructions. The BLEU, CIDEr and ROUGE metrics score human instructions poorly compared to the neural net models.}
\label{tab:metric-scores}
\end{table}

\section{Crafty Details}
\label{sec:appendix_crafty}

We use the data in Matterport3D to build Crafty, a template-based navigation instruction generator that uses a Hidden Markov Model (HMM) to select objects as reference landmarks for wayfinding. Crafty's four main components (\textit{Appraiser}, \textit{Walker}, \textit{Observer} and \textit{Talker}) are described below. 

\subsection{Appraiser}

The Appraiser scores the interestingness of objects based on the Matterport3D scans in the training set.  It treats each panorama as a document and the categories corresponding to objects visible from the panorama as words, and then computes a per-category inverse document frequency (IDF) score. 


\subsection{Walker}

The Walker converts a panorama sequence into a motion sequence. Given a path (sequence of connected panoramas) and an initial heading, it calculates the entry heading into each panorama and the exit heading required to transition to the next panorama. For each panorama, all annotated objects that are visible from the location are retrieved. For each object, we obtain properties such as their category and center, which allows the distance and heading from the panorama center to be computed. From these, the Walker creates a sequence of motion tuples, each of which captures the context of the source panorama and the goal panorama, along with the heading to move from source to goal.

\subsection{Observer}

The Observer selects an object sequence by generating objects from an HMM that is specially constructed for each environment, characterized by:

\begin{itemize}

\item \textit{Emissions}: how panoramas relate to objects. This is a probability distribution over panoramas for each object, based on the distance between the object and the panoramas.

\item \textit{Transitions}: how looking at one object might shift to another one, based on their relative location, the motion at play, and the Appraiser’s assessment of their prominence. 

\end{itemize}

\noindent
The intuition for using an HMM is that we tend to fixate on a given salient object over several steps as we move (high self-transitions); these tend to be nearby (high emission probability for objects near a panorama's center) and connected to the next salient object (biased object-object transitions). To explain a particular observed panorama sequence (path), we can then infer the optimal object sequence using the Viterbi algorithm.

\subsection{Talker}

Given a motions sequence from the Walker and corresponding object observations from the Observer, the Talker uses a small set of templates to create English instructions for each step. We decompose this into low-level and high-level templates.

\subsubsection{Low-level templates}

For single step actions, there are three main things to mention: movement, the fixated object and its relationship to the agent's position. 

\begin{figure}
\begin{center}
\includegraphics[width=1\linewidth]{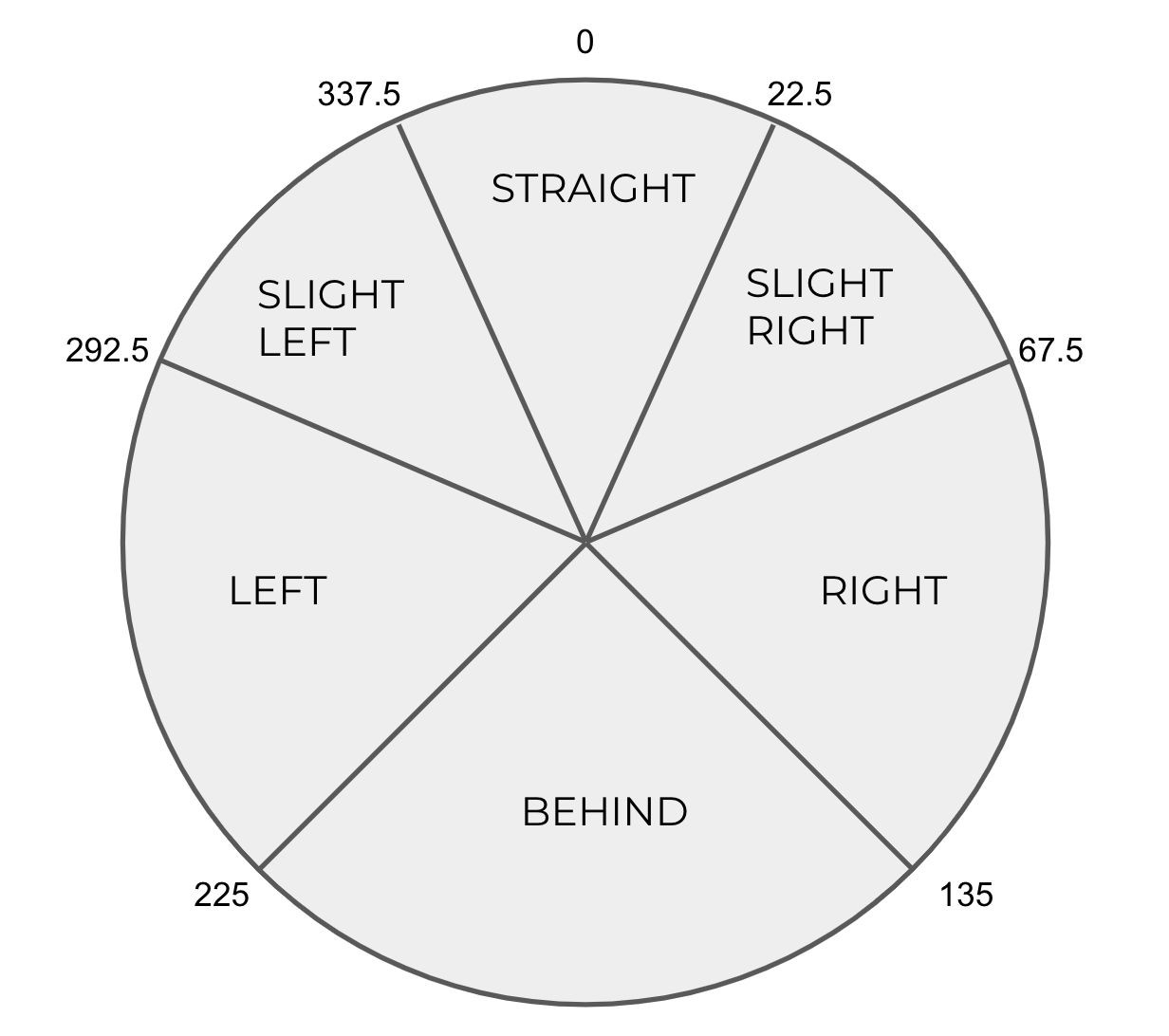}
\end{center}
\caption{Orientation wheel with directions types. The demarcations are $\frac{\pi}{8}$ (22.5$^\circ$), $\frac{3\pi}{8}$ (67.5$^\circ$), etc.}
\label{fig:orientation wheel}
\end{figure}

\textbf{MOVE}. For movement, we simply generate a set of possible commands for each direction type, where the direction types are defined as in the orientation wheel shown in Fig. \ref{fig:orientation wheel}. There are additional direction types for UP and DOWN based on relative pitch (e.g. when the goal panorama is higher or lower than the source). 

Given one of these heading types, we generate a set of matching phrases appropriate to each. E.g. for LEFT and RIGHT, the verbs {\em face}, {\em go}, {\em head}, {\em make a}, {\em pivot}, {\em turn}, and {\em walk} are combined {\em left} and {\em right}, respectively. For moving STRAIGHT, the verbs {\em continue}, {\em go}, {\em head}, {\em proceed}, {\em walk}, and {\em travel} are combined with {\em straight} and {\em ahead}. To select an instruction for a given heading, we randomly sample from the list of available phrases. To generate a MOVE command for LEFT, we randomly sample from [{\em face left}, {\em go left}, $\ldots$ {\em walk left}].

\textbf{OBJ}. An object's description is its category (e.g. {\em couch}, {\em tv}, {\em window}).

\textbf{ORIENT}. We use the same direction types shown in Figure \ref{fig:orientation wheel}. When an object is STRAIGHT and BEHIND, we use the phrases \textit{ahead of you} or \textit{in front of you} and \textit{behind you} or \textit{in back of you}, respectively. For objects to the LEFT or RIGHT, we use two templates \texttt{DIRECTION\_PRE DIRECTION} and \texttt{DIRECTION DIRECTION\_POST}, where \texttt{DIRECTION\_PRE} is selected from [\textit{to your}, \textit{to the}, \textit{on your}, \textit{on the}] and \texttt{DIRECTION\_POST} is the phrase \textit{of you}. This produces \textit{to your left}, \textit{on the right}, \textit{right of you}, and so on. For SLIGHT LEFT and SLIGHT RIGHT, one of [\textit{a bit}, \textit{slightly}, \textit{a little}, \textit{just}] is added in front (e.g. \textit{a bit to your left}).

\subsubsection{High level templates}

Crafty pieces these low-level textual building blocks together to describe actions. In what follows, MOVE, OBJ, and ORIENT indicate the move command, object phrase and orientation phrase, respectively, discussed above.

\textbf{Single action}. We use templates for three situations: start of a path, heading change in a panorama (intra) and moving between panoramas (inter).

\begin{itemize}
    \item \textit{Start of path}: There are several templates that simply help a wayfinder verify their current position. Ex: \textit{you are near a \texttt{OBJ}, \texttt{ORIENT}}.
    \item \textit{Intra}: These templates include the movement command followed by a verification of the orientation to an object having completed the movement. Ex: \textit{\texttt{MOVE}. a \texttt{OBJ} is \texttt{ORIENT}}.
    \item \textit{Inter }: These templates capture walking from one panorama to another and provide additional object verification. Ex: \textit{\texttt{MOVE}, going along to the \texttt{OBJ ORIENT}}.
\end{itemize} 

\textbf{Multi-step actions}. We attempt to reduce verbosity by collapsing actions that involve fixation on the same object. 

\begin{itemize}
    \item \textit{Combining actions}: Repeated actions are collapsed; e.g. [STRAIGHT, STRAIGHT, RIGHT, STRAIGHT] becomes [STRAIGHT, RIGHT, STRAIGHT]). These produce a composite move command, e.g. \textit{proceed forward and make a right and go straight}.
    
    \item \textit{Describing the object}: To orient with respect to the fixated-upon object, we switch on the direction type between the agent and the object at the last action. Ex: for STRAIGHT, we use \textit{heading toward the \texttt{OBJ}} and for SLIGHT LEFT/RIGHT, we use \textit{approaching the \texttt{OBJ ORIENT}}.
\end{itemize}

The final output is the concatenation of the combined move command and the object orientation phrase.

\textbf{End-of-path instruction templates}. The final action is a special situation in that it needs to describing stopping near a salient object. For this, we extract MOVE and OBJ phrases from the last action and use templates such as \textit{\texttt{MOVE} and stop by the \texttt{OBJ}}.

\textbf{Full example}. Putting it all together, Crafty creates full path instructions such as the following, with relevant high-level templates indicated:

\begin{itemize}
    \item (\textsc{Start}) \textit{there is a lamp when you look a bit to the left. pivot right, so it is in back of you.}
    \item (\textsc{Inter}) \textit{walk forward, going along to the curtain in front of you.}
    \item (\textsc{Intra}) \textit{curve left. you should see a tv ahead of you.}
    \item (\textsc{Multi-Action}) \textit{go forward and go slightly left and walk straight, passing the curtain to your right.}
    \item (\textsc{End-of-Path}) \textit{continue forward and stop by the couch.}
\end{itemize}

\crafty's instructions are more verbose than human instructions, but are often easy to follow---provided there are good, visually salient landmarks in the environment to use for orientation.

\section{Human Rater Performance Over Time}
\label{sec:appendix_human_eval}

Human raters are excellent at learning and adapting to new problems over time.  To understand whether our 37 human raters learn to self-correct the perturbed instructions over time and whether that affects the quality of our human wayfinding results, we investigate rater performance as a function of time using the sequence of examples they evaluate.

\begin{figure*}
\centering
\includegraphics[width=0.45\textwidth]{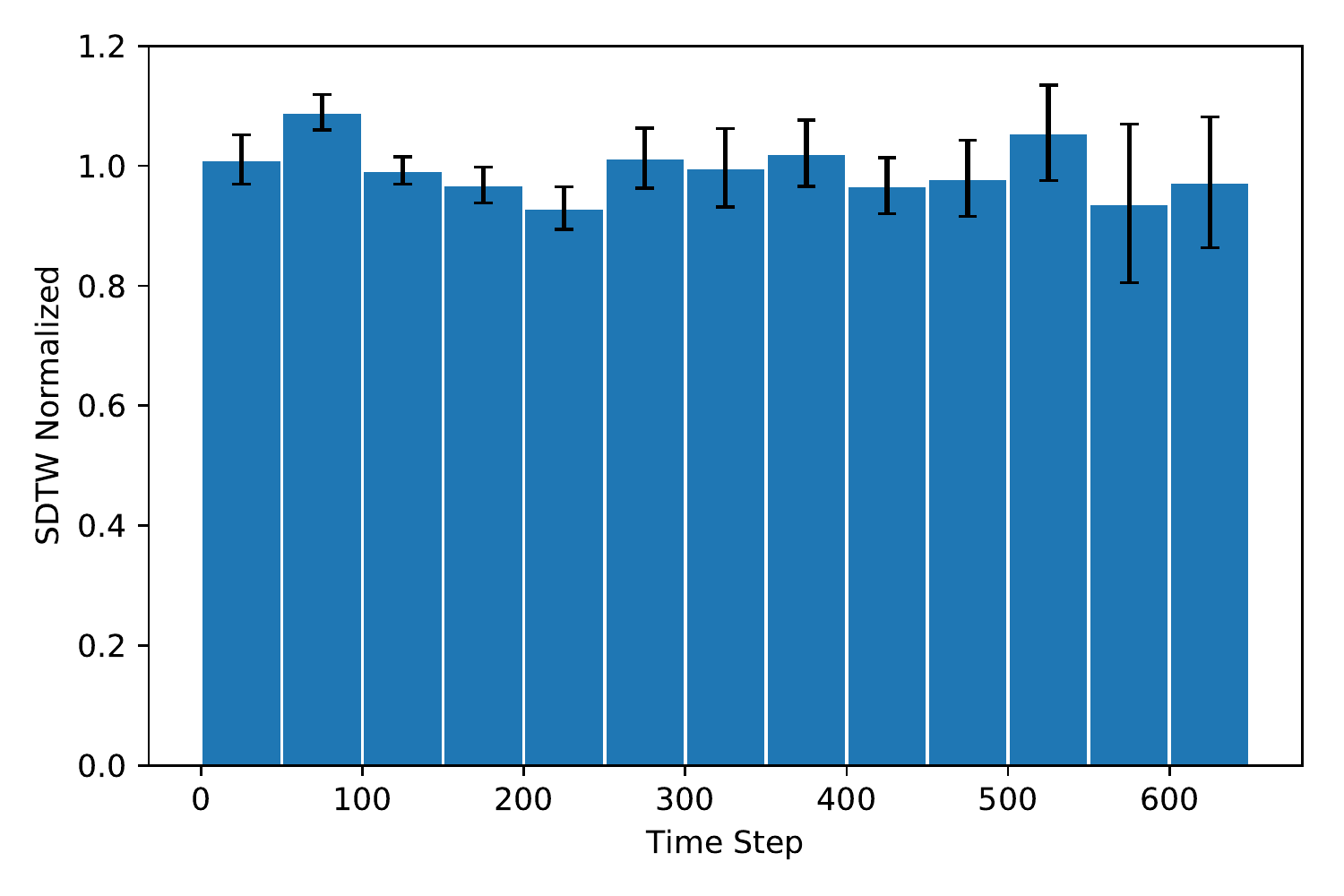}
\includegraphics[width=0.45\textwidth]{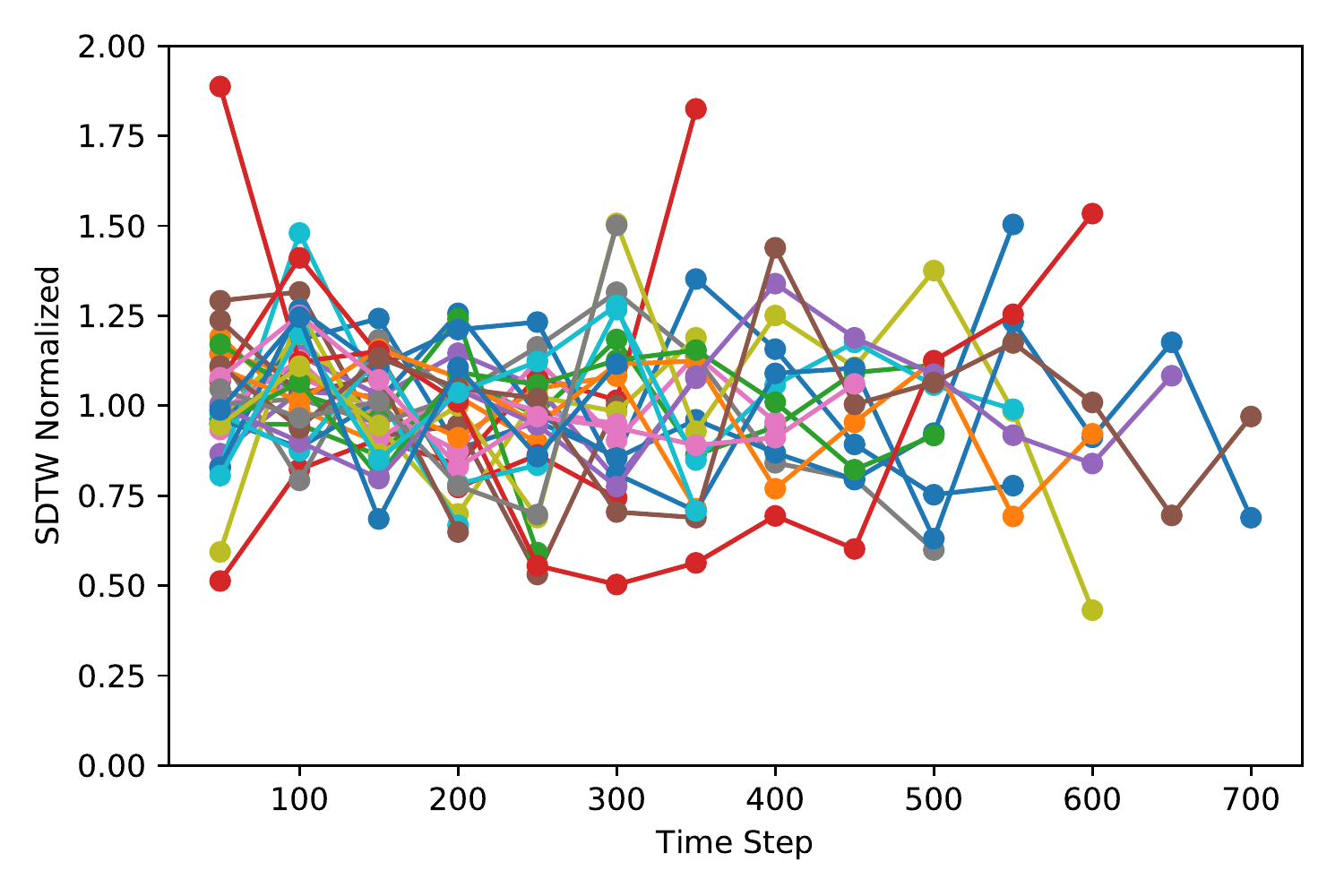}
\caption{Average human rater performance for all datasets as a function of time using the sequence of examples each rater has evaluated. Each rated example indicates a time step. We normalize the  scores of each rater by their mean value over time to remove  performance bias of each rater in order to better pick up the trend over time. We average 50 examples to get the mean SDTW for each rater due to the discrete nature of success. 
\textit{Left}: The mean performance of all rater for each bin.  Error bars represent the standard deviation of the mean. 
\textit{Right}: Individual rater performance over time. Each line represents a single rater. 
Despite a few outliers, the overall human rater performance is flat and consistent over time, indicating no self-correction or adaptation to the datasets by human raters.
}
\label{fig:all_human_rater}
\end{figure*}

\begin{figure*}
\centering
\includegraphics[width=0.328\textwidth]{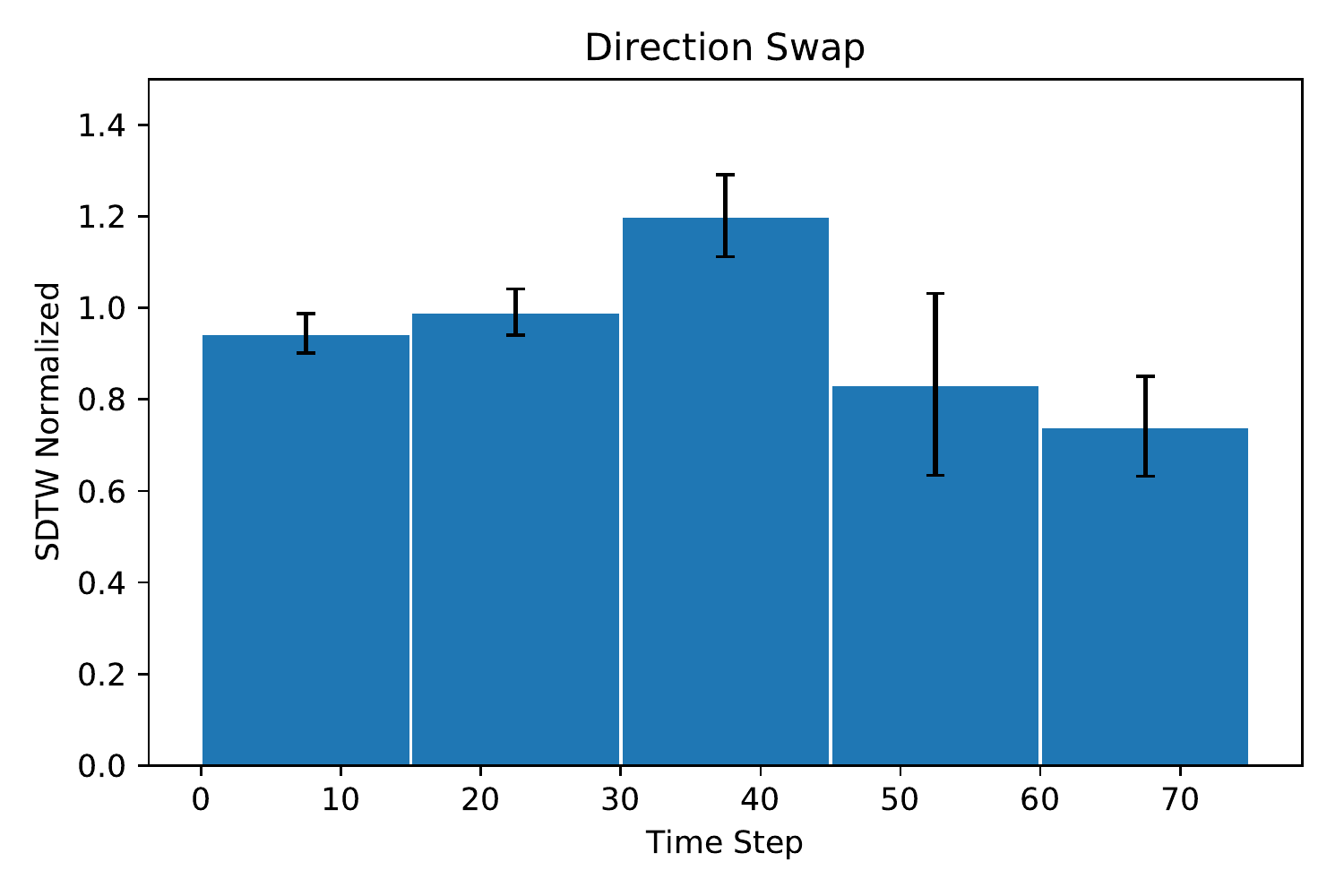}
\includegraphics[width=0.328\textwidth]{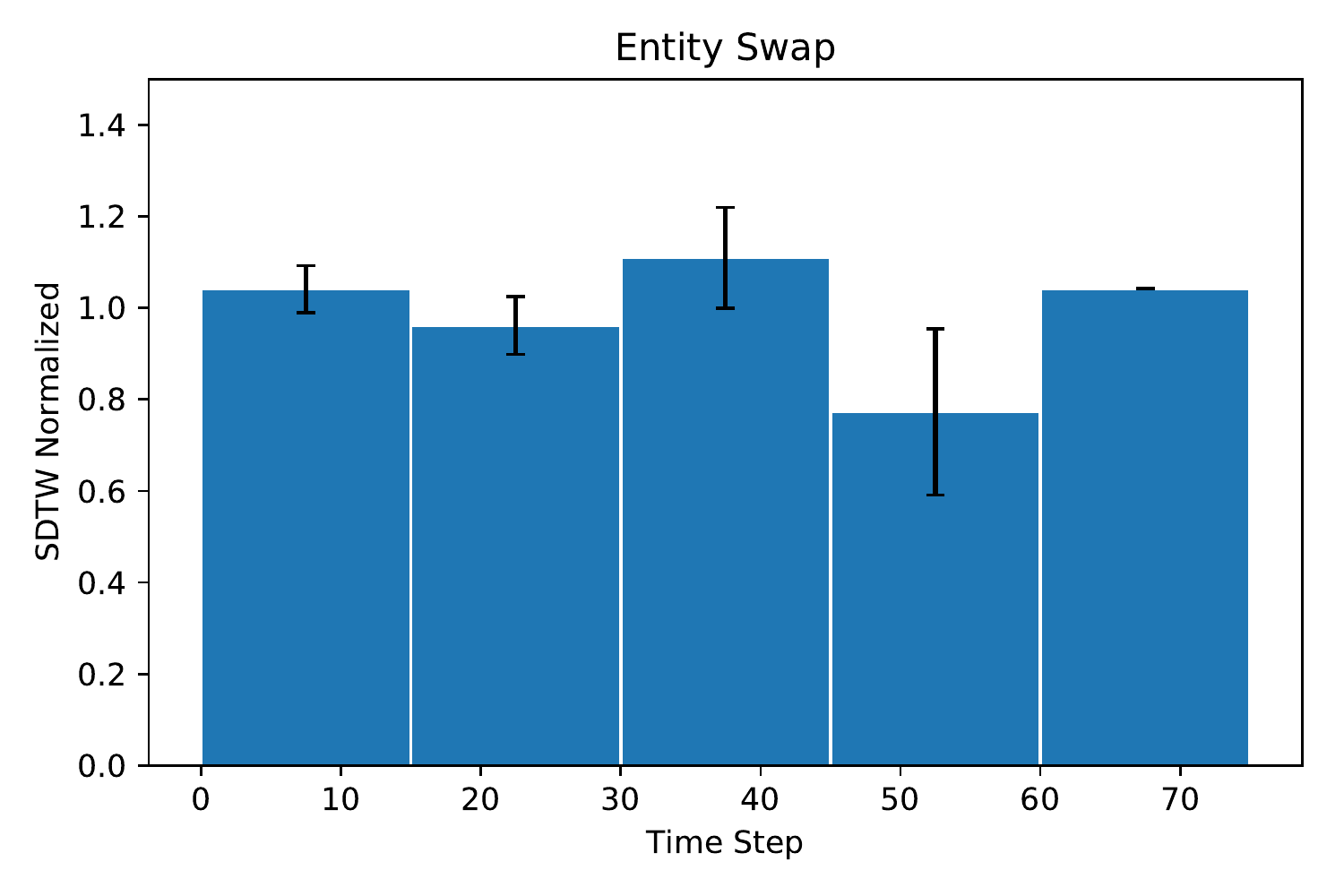}
\includegraphics[width=0.328\textwidth]{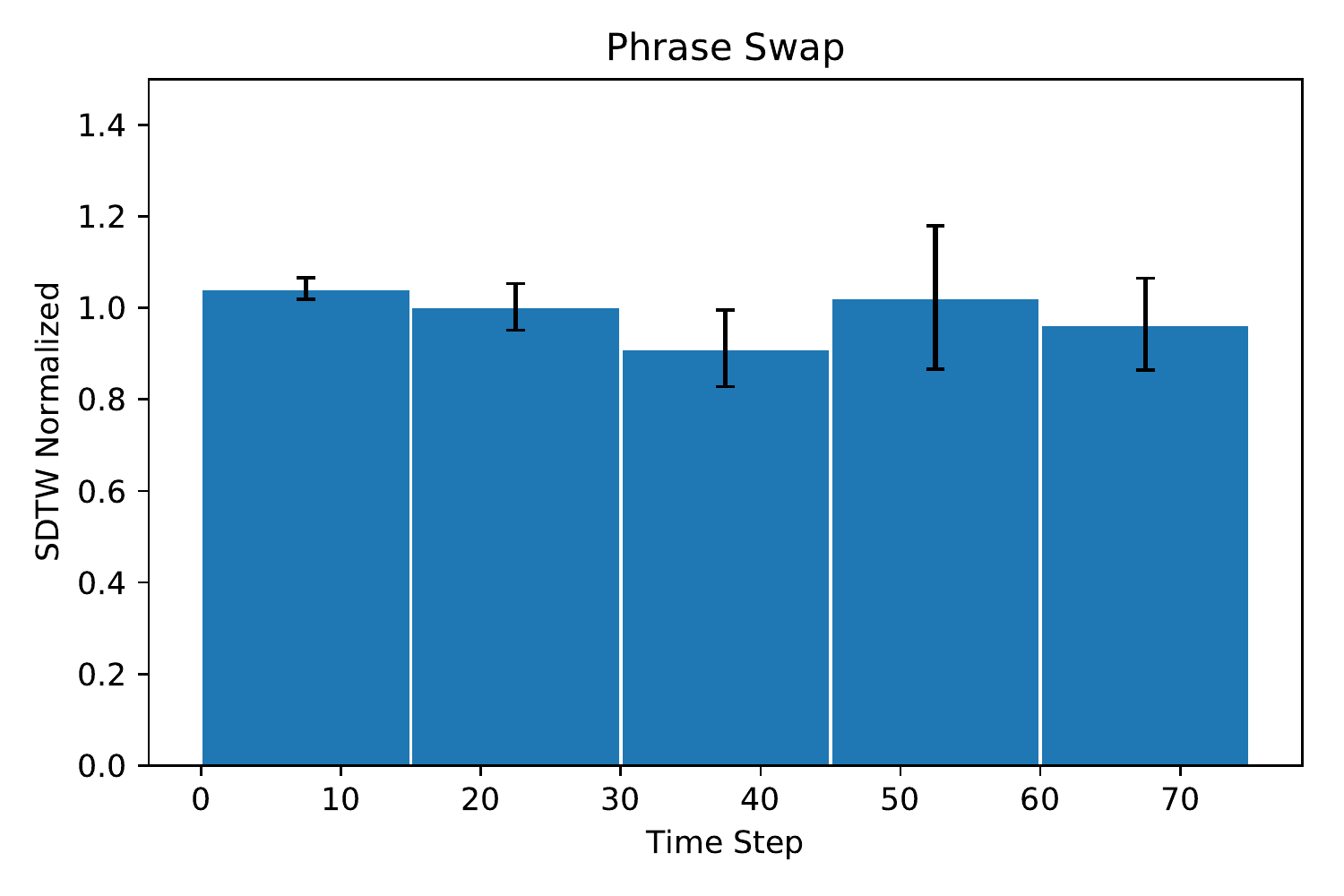}
\caption{Average human rater performance for the three instruction perturbation methods as a function of time (number of examples), computed in a similar way as in Figure \ref{fig:all_human_rater}. We use a 15-point bin to compute the average for each human rater, and aggregate over all raters to get the mean and its uncertainty. The overall human rater performance stays flat and does not drift significantly over time for instruction perturbations.
}
\label{fig:individual_human_rater}
\end{figure*}

Figure \ref{fig:all_human_rater} shows the average human rater performance for all of the 9 datasets included in Table 1 of Section 3. 
Due to the binary nature of SR, we use a 50-point bin to average each rater's performance, and then average the results across all raters for each bin. Figure \ref{fig:all_human_rater} shows that the average rater performance stays flat within the uncertainties and does not show systematic drift over time, indicating no overall self-correction that affects the wayfinding results. For a more granular scrutiny of individual perturbation methods, in particular the perturbed instructions, we plot in Figure \ref{fig:individual_human_rater} the average human rater performance over time for the three methods: \textit{Direction Swap}, \textit{Entity Swap}, and \textit{Phrase Swap}. Despite greater uncertainties due to much fewer data points used for averaging, the overall human performance for each method still does not drift significantly in a systematic manner. These results indicate that our human wayfinding performance results are reliable and robust over time, which can be attributed to shuffling of the examples and that the perturbation methods are blind to human raters.

\end{document}